\documentclass[conference,compsoc]{IEEEtran}
\usepackage{amsmath,graphicx}
\usepackage{multirow}
\usepackage{placeins}
\usepackage[font=small,skip=0pt]{caption}

\ifCLASSOPTIONcompsoc
  \usepackage[nocompress]{cite}
\else
  \usepackage{cite}
\fi

\ifCLASSINFOpdf
\else
\fi

\begin{document}
	\nocite{*}

\title{Sentence-Level Sign Language Recognition Framework}


\author{\IEEEauthorblockN{Atra Akandeh}
\IEEEauthorblockA{Circuits, Systems, and Neural Networks (CSANN) Laboratory \\
Computer Science and Engineering \\
Michigan State University\\
East Lansing, Michigan 48824\\
akandeha@msu.edu}
}

\maketitle

\begin{abstract}

We present two solutions to sentence-level SLR. Sentence-level SLR required mapping videos of sign language sentences to sequences of gloss labels. Connectionist Temporal Classification (CTC) has been used as the classifier level of both models. CTC is used to avoid pre-segmenting the sentences into individual words. The first model is an LRCN-based model, and the second model is a Multi-Cue Network. LRCN is a model in which a CNN as a feature extractor is applied to each frame before feeding them into an LSTM. In the first approach, no prior knowledge has been leveraged. Raw frames are fed into an 18-layer LRCN with a CTC on top. In the second approach, three main characteristics (hand shape, hand position, and hand movement information) associated with each sign have been extracted using Mediapipe. 2D landmarks of hand shape have been used to create the skeleton of the hands and then are fed to a CONV-LSTM model. Hand locations and hand positions as relative distance to head are fed to separate LSTMs. All three sources of information have been then integrated into a Multi-Cue network with a CTC classification layer. We evaluated the performance of proposed models on RWTH-PHOENIX-Weather. After performing an excessive search on model hyper-parameters such as the number of feature maps, input size, batch size, sequence length, LSTM memory cell, regularization, and dropout, we were able to achieve 35 Word Error Rate (WER).

\end{abstract}
 

%
\IEEEpeerreviewmaketitle

\section{Introduction}

According to the World Health Organization (WHO, 2017), 5\% of the world's population have hearing loss. Most people with hearing disabilities communicate via sign language, which hearing people find extremely difficult to understand. To facilitate communication of deaf and hard of hearing people,  developing an efficient communication system is a necessity. There are many challenges associated with the Sign Language Recognition (SLR) task, namely, lighting conditions, complex background, signee body postures, camera position, occlusion, large variations in hand posture, no word alignment, coarticulation, etc. Sign Language Recognition has been an active domain of research since the early 90s. However, due to computational resources and sensing technology constraints, limited advancement has been achieved over the years. 

In this work we study continuous sign language recognition (CSLR) in which temporal boundaries of the words in the sentences are not defined. Most existing CSLR frameworks are only capable of recognizing one sign at a time and require pausing between signs in a sentence. To address this problem, the correspondence between video sequence and sign gloss sequence needs to be learned. Glossing corresponds to mapping signs word-for-word to another written language. Glosses differ from translation as they only denote the meaning of each part in a sign language sentence and do not necessarily construct a grammatically correct sentence in the written language. In this work we do not address linguistic structures and grammar unique to sign language.

This work develops two novel sign language recognition frameworks using deep neural networks which directly map videos of sign language sentences to sequences of gloss labels. The first proposed framework is an 18 layers LRCN-based model in which no prior knowledge is required. The second proposed framework makes use of critical characteristics of the signs and injects domain-specific expert knowledge into the system. This model allows for combining data from variant sources and hence combating limited data resources in the SLR field.

\subsection{Characteristics}

In most Sign Languages, each sign is characterized by manual elements such as shape, movement, and location of the hands. To structure sentences, non-manual elements like eye gaze, mouth shape, facial expression, and body pose are also involved \cite{zhou2020spatial}. 

Facial expressions are used to prevent confusion or misunderstandings. The visible mouth shapes can add information to the meaning of a sign and make it distinguishable from a semantically related sign \cite{kollermouth}. Some ASL signs have a permanent mouth morpheme as part of their production. For example, the ASL word NOT-YET requires a mouth morpheme (TH), whereas LATE has no mouth morpheme. These two are the same signs but with a different non-manual signal \cite{mouthmorphemes}.

In American Sign Language, eye gazing serves a variety of functions. It can regulate turn-taking and mark constituent boundaries. Eye gazing is also frequently used to repair or monitor utterances and to direct the addressee's attention \cite{eyegaze}. It is also engaged in indexing and in expressing object and subject agreement and definiteness versus indefiniteness \cite{thompson2006}.

In this work, non-manual elements, despite their importance, have not been addressed. In the "Future Roadmap" section of this study some suggestions have been made and some references on how to involve those essential factors in recognition systems have been introduced.

\subsection{Challenges}

There are many challenges associated with the SLR task. Publicly available datasets are limited both in quantity and quality. Environmental factors such as lighting sensitivity, complex background, occlusion, signee body postures, and camera position are also challenging issues. In terms of sign linguistics, there are subtle differences between different signs, and there are a large number of vocabularies that need to be learned. Specifically, in sentence-level SLR, there is no word alignment, and temporal boundaries of a specific word are not clear. Also, sentence length variation caused by the number of the word or by signing speed need to be considered. Moreover, signs are context-dependent and, coarticulation in which sign is affected by the preceding or following signs plays an important role. All these factors together make sentence-level SLR a very challenging task.

\subsection{Related Work}

\cite{koller2016deeph} employed a pre-trained 22-layer CNN model within an iterative EM algorithm on a sequence of data. The algorithm iteratively refined the frame-level annotation and subsequent training of the CNN. Three thousand manually labeled hand shape images of 60 different classes were employed to train the model.

\cite{koller2016deeps} embedded a CNN into an HMM. The outputs of the CNN were treated as Bayesian posteriors, and they trained the system in an end-to-end fashion. Their model was very similar to the LRCN mentioned earlier, which combined CNN and LSTM \cite{akandeh2019slim}. They were able to improve the state-of-the-art accuracy on three challenging continuous sign language benchmarks (SIGNUM, RWTH-PHOENIX-Weather2012, RWTH-PHOENIX-Weather Multi-signer) by 15\%, 38\% and 13.3\% respectively.

\cite{cui2017recurrent} used a CNN-LSTM model to learn the mapping of sign sequences to sequences of the gestures on RWTH-PHOENIX-Weather Multi-signer and achieved satisfactory performance. Their architecture consists of a CNN with temporal convolution and spatial pooling, a bidirectional LSTM for global sequence learning, and a detection network. The detection network combines the temporal convolution operations on the spatio-temporal features. According to the authors, this combination acts like a sliding window along with the sign sequences.

To perform the Sign2Gloss task, \cite{camgoz2020sign} utilized a pre-trained Inception model as  spatial embeddings in a CNN+LSTM+HMM setup. They extracted frame level representations from sign videos and trained two-layered sign language transformers to learn CSLR (continuous sign language recognition) and SLT(sign language translation) jointly in an end-to-end manner. \cite{niu2020stochastic} also proposed to use the transformer encoder as the contextual model for CSLR to fine-tune the lower-level visual feature extractor during model training. To improve model robustness, they proposed dropping video frames stochastically (SFD) and randomly stopping the gradients of some frames during training (SGS).

\section{Proposed Models}

We present the development and implementation of two deep sentence-level sign language recognition frameworks. In the first framework, to minimize prior knowledge, raw frames are fed to a deep network. In the second framework, sign language linguistics knowledge is leveraged to capture three main components of each sign. In both models, we take advantage of Connectionist Temporal Classification to overcome pre-segmentation requirements posed by the sequence learning problem associated with sentence-level SLR. We call the first proposed model RSign-C (R corresponds to raw input and C corresponds to CTC), and the second model Multi-Cue Sign-C (MCSign-C).

\subsection{Model Overview}

In this section, we provide an overview of two proposed models, namely, RSign-C and MCSign-C. These models are developed based on word-level SLR architectures that have been proposed in \cite{my-thesis}, except we modify the $softmax$ layer, which calculates the probability of a word and adds a CTC layer that calculates the probabilities of a sentence.

\begin{figure}[!htb]
	\centering
	\includegraphics[trim={0 0 0 0},clip,scale=0.32]{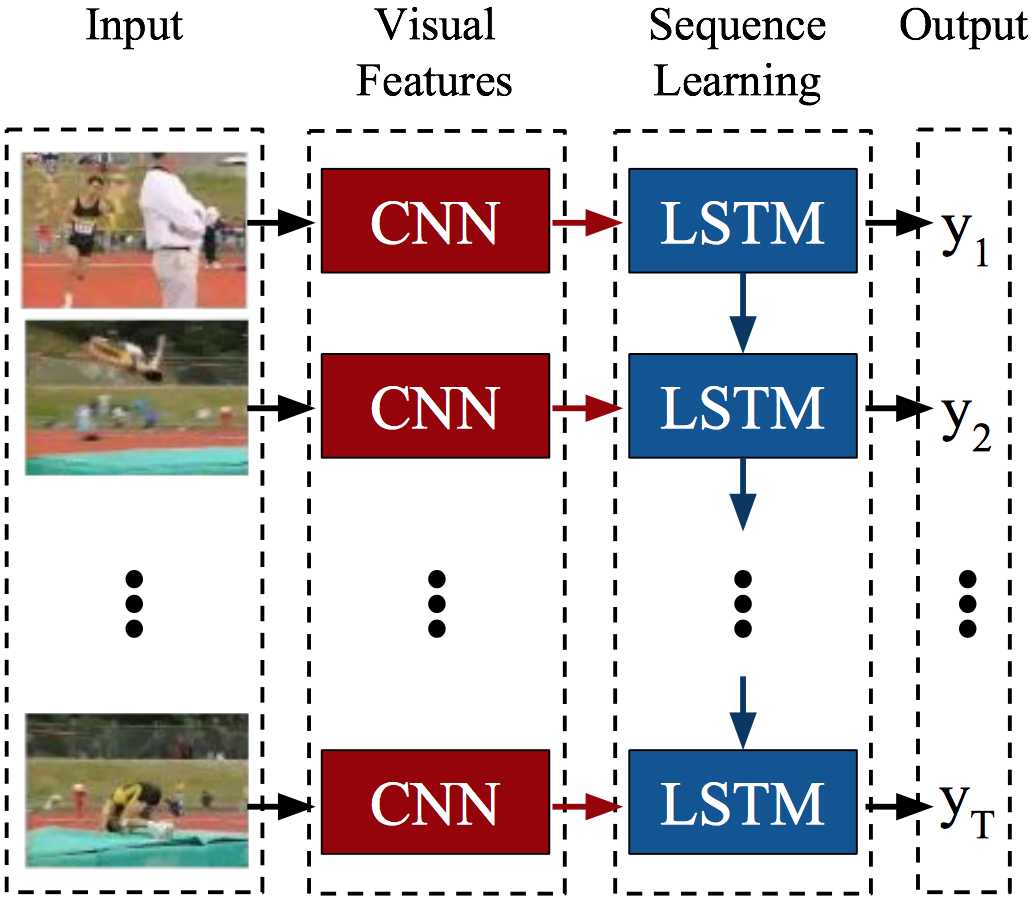}
	\caption{Long-term Recurrent Convolutional Network \cite{Donahue2017}}
	\label{donahue-net}
\end{figure}

RSign-C is an 18-layers LRCN with a CTC layer on top. In the LRCN network, a CNN model as a feature extractor is applied to each frame before feeding them into an LSTM \cite{akandeh2017simplified}. Figure \ref{donahue-net} depicts an overview of an LRCN model.

\begin{figure}[!htb]
	\centering
	\includegraphics[trim={0 0 0 0},clip,scale=0.3]{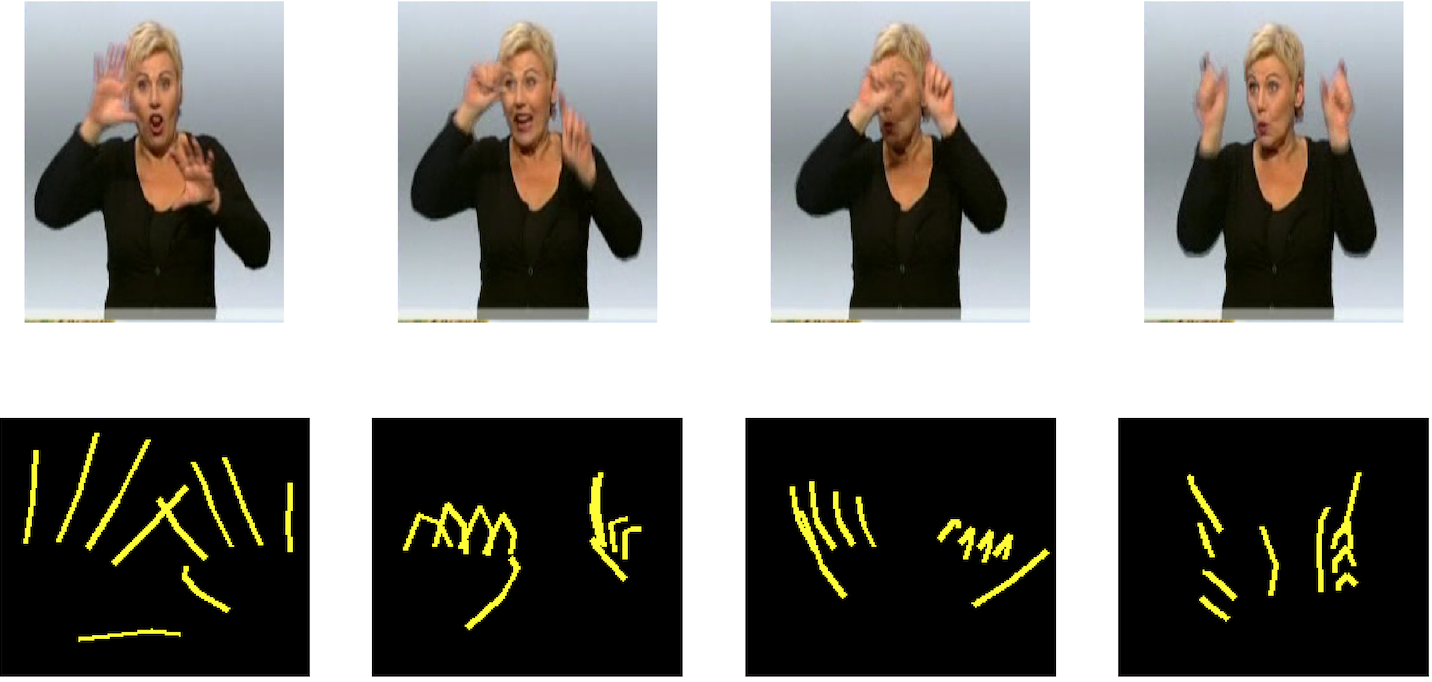}
	\caption{New black \& white image sequence created}
	\label{s-black-white}
\end{figure}

\begin{figure}[!htb]
	\centering
	\includegraphics[trim={0 0 0 0},clip,scale=0.22]{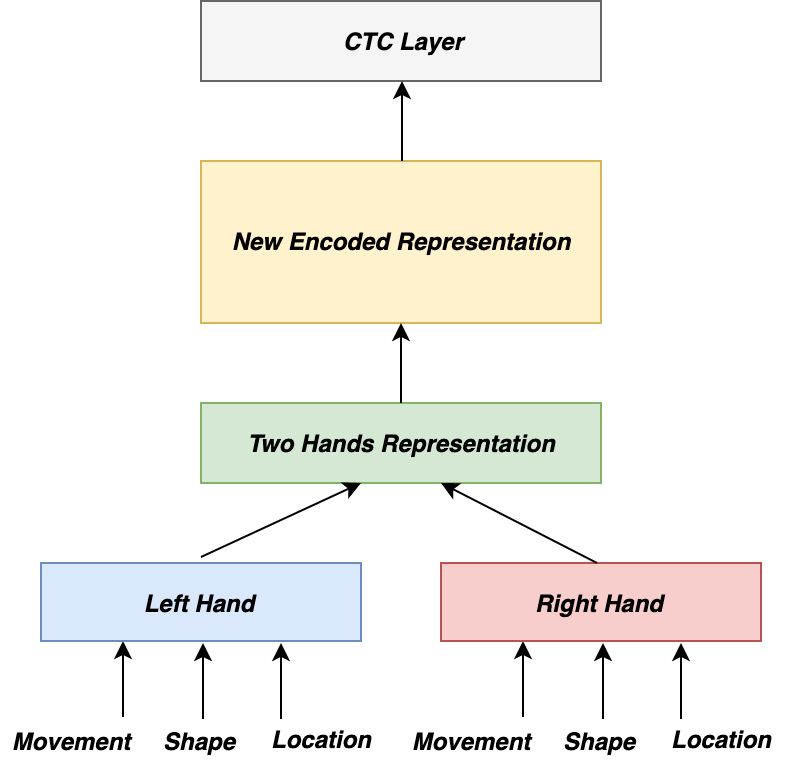}
	\caption{MCSign-C model overview}
	\label{s-model-overview}
\end{figure}

MCSign-C, on the other hand, aims to capture and model the three main characteristics associated with each sign. At the first layer, a sequence of 2D landmarks of hands, as well as 2D landmarks of the upper-body are captured during the signing. At the second layer, as it was suggested by \cite{fang2017deepasl} the critical components of the signs, namely shape, movement, and location of hands are modeled. A new 2D black and white image (see figure \ref{s-black-white}) is created out of 2D landmarks of hands to model the handshapes. Hands movement is encoded by spatial displacement of the palm center between two consecutive frames. A one-hot vector corresponding to the relative location of hands to the head is also created to encode the hands' location. At the third level, newly created spatio-temporal trajectories are fed into a CONV-LSTM model. The one-hot vector corresponding to the hand locations is also fed into an LSTM model. Then, at the last layer, a CTC classifier outputs the sequences of gloss labels. Figure \ref{s-model-overview} provides an overview of the proposed architecture. 

\subsection{Model Details}

This section provides a detailed description of two proposed models, namely, RSign-C and MCSign-C. 

\subsubsection{RSign-C}

\begin{figure}[!htb]
	\centering
	\includegraphics[trim={0 0 0 0},clip,scale=0.16]{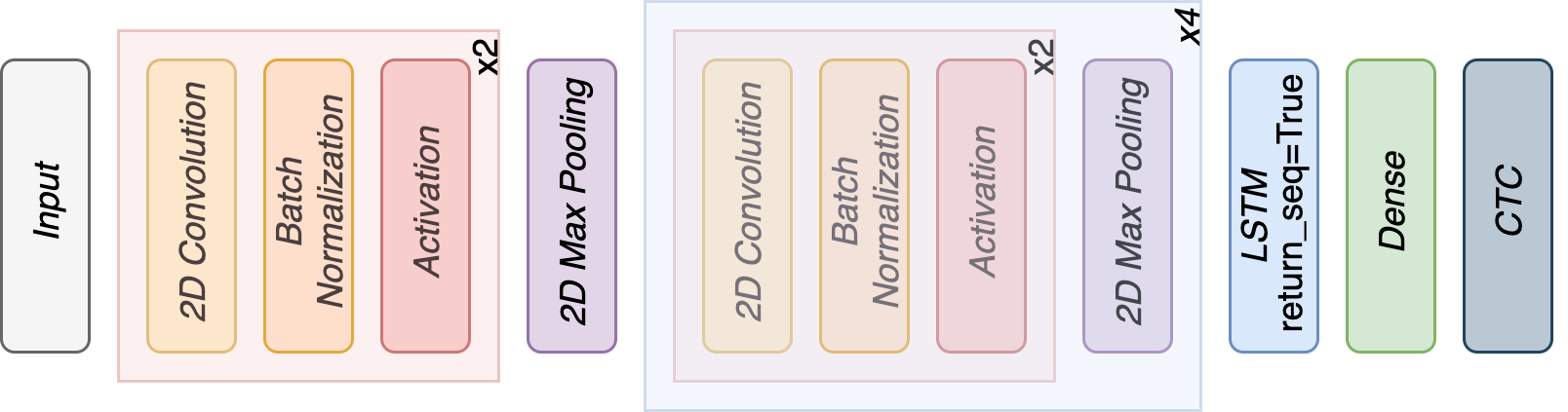}
	\caption{RSign-C model overview}
	\label{s-lrcn}
\end{figure}

We modify the RSign model \cite{my-thesis} such that it is capable of computing the probabilities of a sentence. The LRCN proposed by \cite{Donahue2017} (figure \ref{donahue-net}) has been used as the building block of the RSign model. To produce a probability distribution over all labels at each time step, we modify the final LSTM to return all sequences corresponding to each hidden state, then apply a $softmax$ function and finally add a CTC layer. The detailed architecture used in this study is given in figure \ref{s-lrcn}. As mentioned earlier, we used the Time-Distributed layer in Keras to apply feature extraction to every temporal slice of the input. 

\subsubsection{MCSign-C}

\begin{figure}[!htb]
	\centering
	\includegraphics[trim={0 0 0 0},clip,scale=0.36]{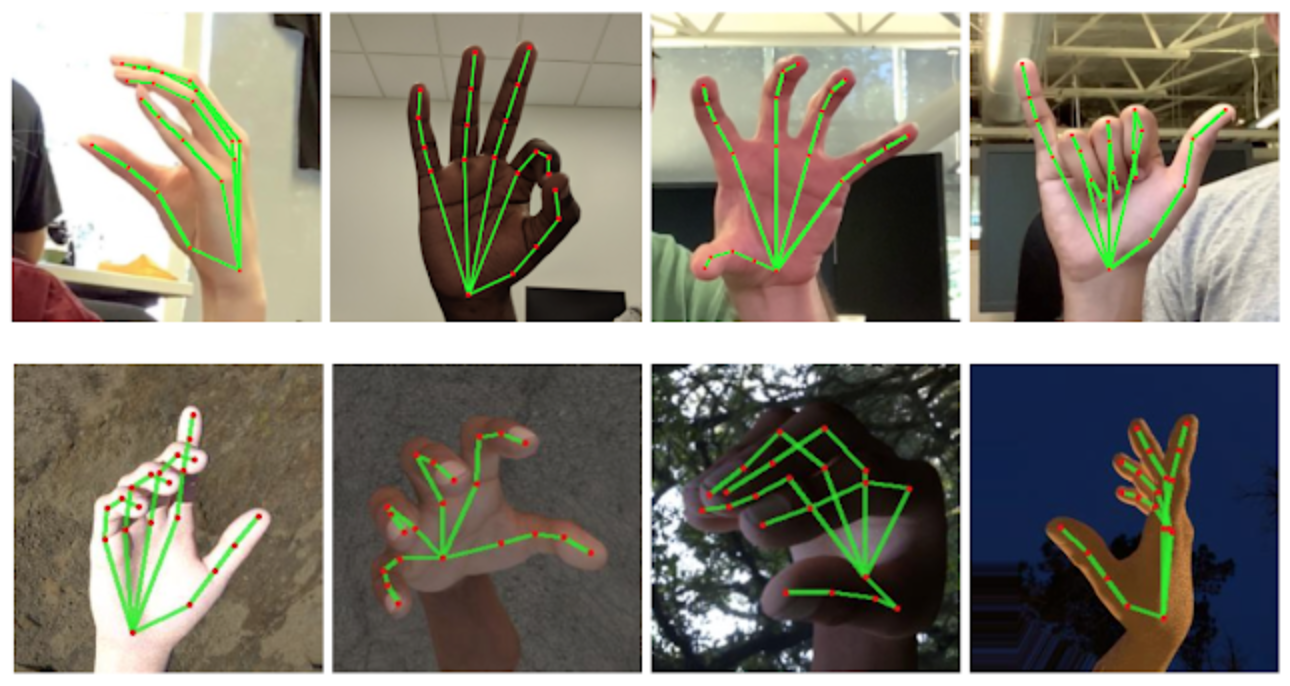}
	\caption{MediaPipe hands tracking module output example  \cite{mediapipe}}
	\label{mp-hands}
\end{figure}

To emphasize on crucial characteristics of the signs and inject domain-specific expert knowledge into the system, we propose to extract and model each characteristic as three input modalities (cues) to the network. Multi-hands tracking and pose modules in Mediapipe \cite{mediapipe} have been used to extract this representative information (see figure  \ref{mp-hands}  for the 2D landmarks of hands tracked by Mediapipe). 

\begin{figure}[!htb]
	\centering
	\includegraphics[trim={0 0 0 0},clip,scale=0.11]{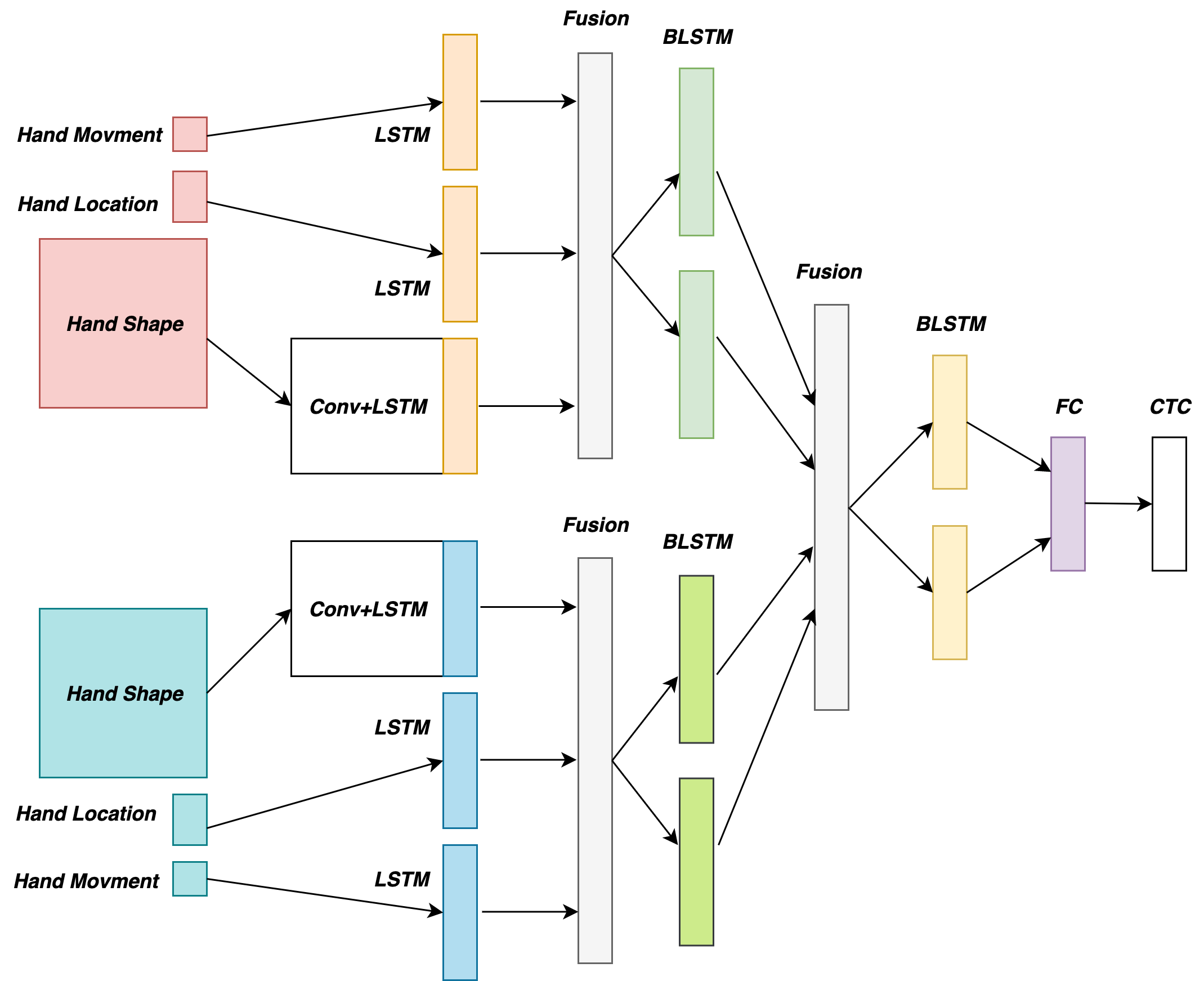}
	\caption{MCSign-C architecture }
	\label{s-model-whole}
\end{figure}

Having access to 21 2D landmarks of the hands, we create new black and white images representing handshape in each frame. We also capture hand movement information of hands by comparing successive frames and calculating palm center displacements. Hand location is also encoded in a one-hot vector based on hand closeness to eyes, mouth, or chest. The extracted characteristics result in six input sequences that represent information related to each handshape, movement, and location. Figure \ref{s-model-whole} illustrates the architecture of the proposed model.

\begin{figure*}[!htb]
	\centering
	\includegraphics[trim={0 0 0 0},clip,scale=0.17]{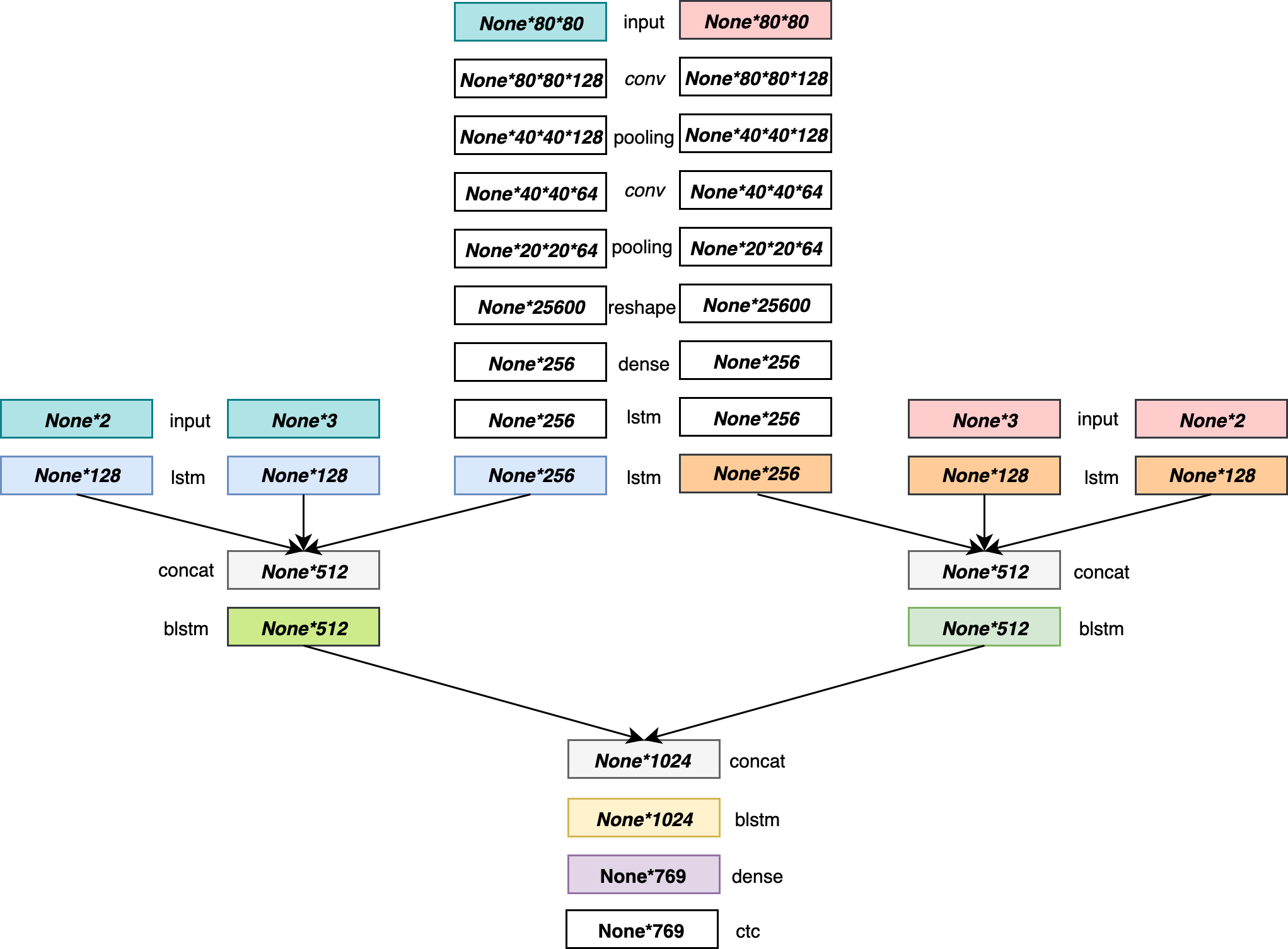}
	\caption{MCSign-C model summary}
	\label{s-dim}
\end{figure*}

As shown in figure \ref{s-model-whole}, our proposed model consists of seven main layers. Within the first layer, a block of CONV-LSTM is embedded. This CONV-LSTM model is also LRCN-based in which a CNN model as a feature extractor is applied to each frame before feeding them into an LSTM. For the CNN part, we employ the SConv model proposed in \cite{my-thesis}. At the first layer, the new skeleton images of each hand are fed to the CONV-LSTM block. Hand movement and hand location are also fed into two separate LSTM. The new representation of each characteristic is then concatenated in the first fusion layer and fed into BLSTM to create a unified representation of each hand. Two high-level hand results are then fused and fed into another BLSTM. Finally, The results of this process are fed into a fully connected layer with CTC loss that drives the final classification decision. See figure \ref{s-dim}, for a detailed model summary. The \textit{'None'} in this figure refers to the number of frames, which can be variable and depends on the signee speed or length of the sentence. 

\subsection{Design Choice}

There have been many different architectures proposed by many researchers to address the SLR task. Here, we present the main reasons behind selecting some specific design choices.

In the MCSign-C model, we leverage sign language linguistics knowledge to capture representative information from the frame sequences. To extract handshape information, 2D landmarks of both hands are outputted. Since there is no ultimate solution on how to lay out the landmarks, we leave it to the network to decide. We regenerate each frame using 2D landmarks as a 2D black and white image and feed it to the network. 

As many signs share similar characteristics at the beginning of their trajectories, to avoid any confusion by traditional unidirectional LSTM, a bidirectional LSTM model has been adopted (also suggested by  \cite{fang2017deepasl}). A bidirectional LSTM model performs inference based on both past and future information and computes the hidden state sequence by combining the output sequences of LSTM by iterating forwards and backward.

To perform sentence-level SLR, a pre-segmented sentence is required as an input to the current technologies. This restriction requires users to pause after signing each word in a sentence. To address this problem, some researchers have proposed sign boundary detection to detect when a sign ends, and the next one begins. Attention mechanisms (Encoder-Decoder Networks) and Connectionist Temporal Classification (CTC)  have been also proposed by other researchers to compute the probability of the whole sentence directly. As CTC can also handle data with different  sequence lengths and it is being used in many natural processing systems, we implement a framework with a CTC classifier to avoid  pre-segmenting the sentences.
 
\section{Dataset}

The RWTH-PHOENIX-Weather \cite{forster2012rwth}, is a dataset for continuous sign language recognition. This dataset contains 5672 sentences in German sign language for training with 65,227 signs and 799,006 frames in total. Table \ref{rwth} summarizes setup statistics \cite{rwth}.

\begin{table}[ht]
	\caption{RWTH-PHOENIX Setup Statistics}
	\centering
	\begin{tabular}{| c | c | c |c|} 
		\cline{3-4}
		 \multicolumn{2}{c|} {} & Glosses & German \\
		\hline
		Train & number of sentences  & 2612 & 2612 \\ 
		\cline{2-4}
		 & number of running words & 20713 & 26585 \\
		\cline{2-4}
		 & vocabulary size & 768 & 1389 \\
		\cline{2-4}
		 & singeltons/vocabulary size & 32.4\% & 36.4\% \\
		\hline
		Development & number of sentences & 250 & 250 \\
		\cline{2-4}
		& number of running words & 2573 & 3293 \\
		\cline{2-4}
		& out-of-vocabulary-words & 1.4\% & 1.9\% \\
		\hline
		Test & number of sentences  & 228 & 228\\ 
		\cline{2-4}
		& number of running words & 2163 & 2980\\
		\cline{2-4}
		& out-of-vocabulary-words & 1.0\% & 1.5\% \\
		\hline
	\end{tabular}
	\label{rwth}
\end{table}

\section{Experimental Setup}

To implement the CNN block of the CONV-LSTM network, we employ the SConv model \cite{my-thesis} with TimeDistributed layer in Keras. The only difference is to perform batch-normalization and then apply $relu$ nonlinearity after each convolutional layer.

As mentioned before, in the CTC layer, a probability distribution over all labels at each time step is predicted. This can be implemented by feeding a 2D input (e.g., the output of an LSTM layer where return-sequence is true) to a Softmax function and then employ the CTC loss function.

To infer a likely output, the Beam Search algorithm has been used. Only a predetermined number of best partial solutions (beam size) are kept as candidates in the beam search algorithm. To get the optimum beam size performing error analysis are required.  If the probability of a true label $y^{*}$ for a given input is lower than the probability of generated output $\tilde{y}$, then the network is at fault, and more training is required. If $p(y^{*})> p(\tilde{y})$,  then the beam search is at fault, and the beam size needs to be increased.

\section{Experimental Results}

To evaluate the performance of the two proposed models, we use word error rate (WER) as the evaluation metric. WER  is a standard metric of the performance of a sentence-level translation or a speech recognition system and is computed as the required number of insertions, substitutions, and deletions to get from the reference to the hypothesis, divided by the total number of words in the reference. 

After performing an excessive search on model hyper-parameters such as the number of feature maps, input size, batch size, sequence length, LSTM memory cell, regularization, and dropout MCSign-C achieves 35.2\% word error rate on mapping Dev sentences and 35.3\% word error rate on Test sentences. RSign-C also  achieves 45.1\% word error rate on mapping Dev sentences and 45.1\% word error rate on Test sentences. Table \ref{wer} lists other approaches and their corresponding WER reported on RWTH-PHOENIX-Weather dataset.

\begin{table}[ht]
	\caption{ Comparison between various approaches on the RWTH-PHOENIX dataset}
	\centering
	\begin{tabular}{| c | c | c |} 
		\hline
		Method & Dev(WER) & Test(WER) \\
		\hline
		CMLLR \cite{koller2015continuous} & 55.0  & 53.0 \\ 
		1-Mio-Hands \cite{koller2016deeps} & 45.1 & 47.1 \\
		CNN-Hybrid \cite{koller2016deeps} & 38.3 & 38.8 \\
		SubUNets \cite{camgoz2017subunets} & 40.8 & 40.7 \\
		Staged-Opt \cite{cui2017recurrent}& 39.4 & 38.7 \\
		Re-sign \cite{koller2017re} & $27.1$ & 26.8 \\
		LS-HAN \cite{huang2018video} & $ - $  & 38.3 \\
		Dilated  \cite{pu2018dilated} & 38.0 & 37.3 \\
		Hybrid CNN-HMM \cite{koller2018deep} & 31.6 & 32.5\\
		IAN \cite{pu2019iterative} & 37.1  & 36.7\\
		DenseTCN \cite{guo2019dense} & 35.9  & 36.5\\
		DNF \cite{cui2019a} & 23.8  & 24.4\\
		DNF \cite{cui2019a} & 23.1  & 22.9\\
		CNN-LSTM-HMM \cite{koller2020weakly} & 26.0  & 26.0\\
		\hline
		\hline
		RSign-C & 45.1 & 45.1 \\
		MCSign-C & 35.2 & 35.3 \\
		\hline
	\end{tabular}
	\label{wer}
\end{table}

\section{Discussion}

With the second proposed model, we are able to achieve 35 WER on WTH-PHOENIX-Weather dataset. The slightly lower performance of the model is compensated by greater robustness to environmental circumstances. By creating black and white skeleton images using Mediapipe from raw data, we not only overcome lighting conditions, complex background, and camera position challenges but also create a way to easily combine data from variant sources and hence combat limited data resources in the SLR field. Since the proposed network is insensitive to dataset type, combining different datasets boosts the network's performance tremendously. 

\section{Future Roadmap }

Although this study aimed to address many aspects of SLR that computer scientists have overlooked, there are still significant gaps to be addressed. Specifically, non-manual elements, including the eye gaze, mouth shape, facial expression require much attention in SLR. Face Mesh and Iris tracking modules in MediaPipe can be beneficial to pursue this track.

To extend the SLR task, one may also look into SLT. SLT differs from SLR as the latter merely detects a sequence of signs without taking into account the linguistic structures and grammar unique to sign language \cite{yin2020sign}. RWTH-PHOENIX-Weather is currently the only publicly available dataset with both gloss labels and spoken language translations. 

There have been limited studies on mapping glosses to spoken language. To map the detected glosses into a proper sentence in the target language Neural Machine Translation (NMT) has been introduced. An encoder-decoder architecture, also known as a sequence to sequence model, is primarily employed in recent NMT approaches. However, sequence to sequence networks are unable to model long-term dependencies in large input sentences. To address this issue, attention mechanisms are introduced  \cite{bahdanau2015neural}. Transformer \cite{vaswani2017attention} is an encoder-decoder network in which  self-attention layers are used in place of recurrent networks. The next phase in translating sign language to written or spoken language can be piping the output of an SLR system into a sequence to sequence model. 

\section{Conclusion}

In this study, we proposed two solutions to sentence-level SLR. Connectionist Temporal Classification (CTC) has been used as the classifier level of both models. In the first approach, no prior knowledge has been leveraged. Raw frames are fed into an 18-layer LRCN with a CTC on top. In the second approach, three main characteristics (hand shape, hand position, and hand movement information) associated with each sign have been extracted using Mediapipe. 2D landmarks of handshape have been used to create the skeleton of the hands and then are fed to a CONV-LSTM model. Hand locations and hand positions as relative distance to head are fed to separate LSTMs. All three sources of information have been then integrated into a Multi-Cue network with a CTC classification layer. We evaluated the performance of proposed models on RWTH-PHOENIX-Weather and were able to achieve 35 Word Error Rate.

\small{
	\bibliographystyle{ieee}
	\bibliography{egbib}
}

\end{document}